\documentclass[letterpaper]{article} 
\usepackage{aaai2026}  
\usepackage{times}  
\usepackage{helvet}  
\usepackage{courier}  
\usepackage[hyphens]{url}  
\usepackage{graphicx} 
\usepackage{natbib}  
\usepackage{caption} 
\usepackage{algorithm}
\usepackage{algorithmic}
\usepackage{times}
\usepackage{soul}
\usepackage{url}
\usepackage{caption}
\usepackage{amsmath}
\usepackage{amsthm}

\usepackage{amssymb}
\usepackage{booktabs}
\usepackage{makecell}
\usepackage{multirow}
\usepackage{algorithm}
\usepackage{algorithmic}
\usepackage{lineno}
\usepackage{xcolor}
\definecolor{myred}{HTML}{FF2600}
\definecolor{mypurple}{HTML}{7030A0}
\definecolor{myblue}{HTML}{0096FF}

\newcommand{\ssymbol}[1]{^{\@fnsymbol{#1}}}

\usepackage{newfloat}
\usepackage{listings}
\DeclareCaptionStyle{ruled}{labelfont=normalfont,labelsep=colon,strut=off} 
\floatstyle{ruled}
\newfloat{listing}{tb}{lst}{}
\floatname{listing}{Listing}
\title{UniCUE: Unified Recognition and Generation Framework for Chinese Cued Speech Video-to-Speech Generation}
\author{
    Jinting Wang\textsuperscript{\rm 1},
    Shan Yang\textsuperscript{\rm 2},
    Chenxing Li\textsuperscript{\rm 2},
    Yu Dong\textsuperscript{\rm 2},
    Li Liu\textsuperscript{\rm 1}\footnote{Corresponds to Li Liu (avrillliu@hkust-gz.edu.cn)}
}
\affiliations{
    \textsuperscript{\rm 1}The Hong Kong University of Science and Technology (Guangzhou)\\

     \textsuperscript{\rm 2}Tencent AI Lab

}

\usepackage{bibentry}

\begin{document}

\maketitle

\begin{abstract}
Cued Speech (CS) enhances lipreading via hand coding, offering visual phonemic cues that support precise speech perception for the hearing-impaired. The task of \textbf{CS V}ideo-to-\textbf{S}peech generation (\textbf{CSV2S}) aims to convert CS videos into intelligible speech signals.
Most existing research focuses on \textbf{CS R}ecognition (\textbf{CSR}), which transcribes video content into text. Consequently, a common solution for CSV2S is to integrate CSR with a text-to-speech (TTS) system. However, this pipeline relies on text as an intermediate medium, which may lead to error propagation and temporal misalignment between speech and CS video dynamics. In contrast, directly generating audio speech from CS video (direct CSV2S) often suffer from the inherent multimodal complexity and the limited availability of CS data.
To address these challenges, we propose \textbf{UniCUE},
the first unified framework for CSV2S that \textbf{directly} generates speech from CS videos without relying on intermediate text.
The core innovation of UniCUE lies in integrating a understanding task (CSR) that provides fine-grained CS visual-semantic cues to to guide the speech generation.
Specifically, UniCUE incorporates a pose-aware visual processor, a semantic alignment pool that enables precise visual–semantic mapping, and a VisioPhonetic adapter to bridge the understanding and generation tasks within a unified architecture.
To support this framework, we construct UniCUE-HI, a large-scale Mandarin CS dataset containing 11,282 videos from 14 cuers, including both hearing-impaired and normal-hearing individuals. 
Extensive experiments conducted on this dataset demonstrate that UniCUE achieves state-of-the-art (SOTA) performance across multiple evaluation metrics. Project website can be found at \url{https://beria-moon.github.io/UniCUE/}
\end{abstract}



\begin{figure}[t]
    \centering
    \includegraphics[width=0.99\linewidth]{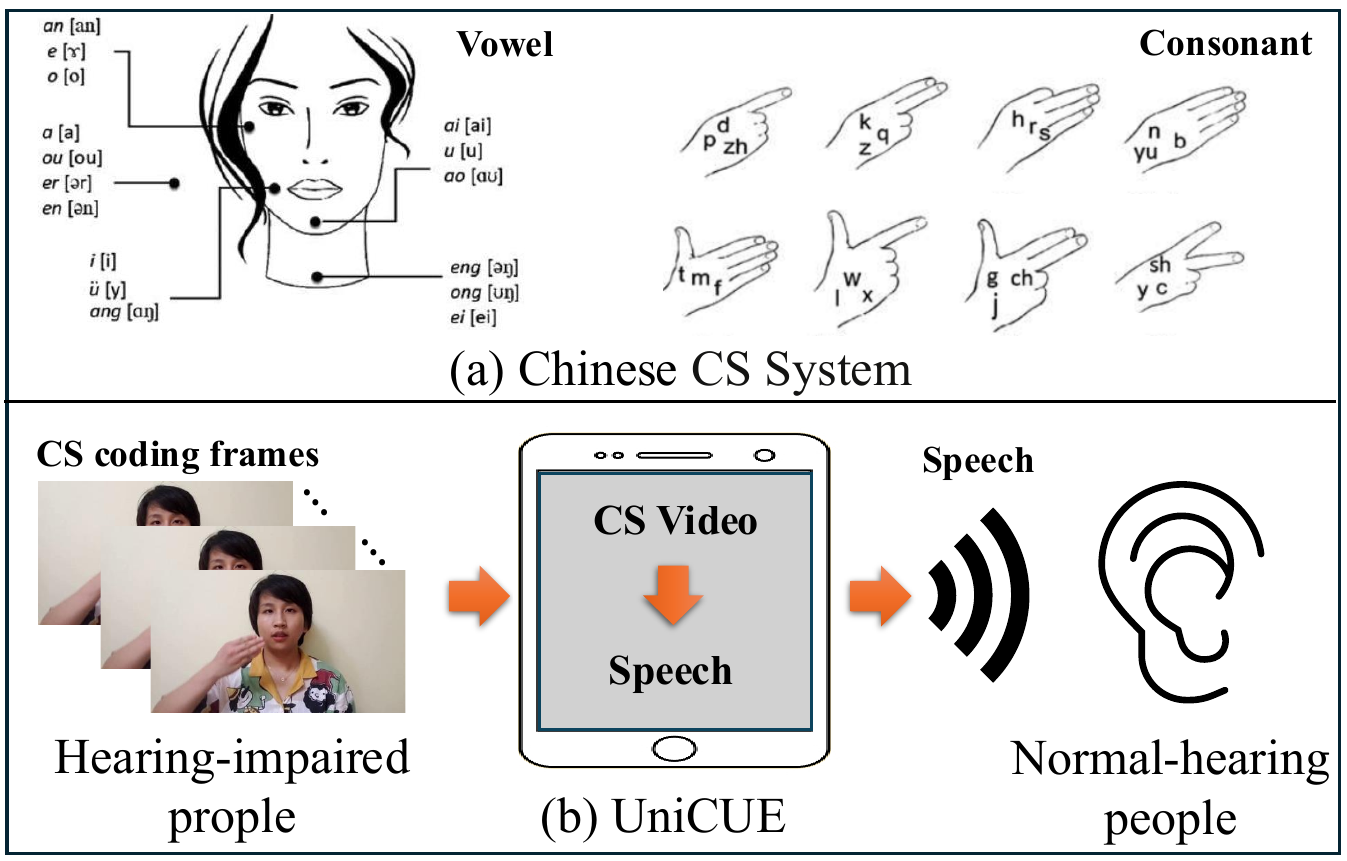}
    \caption{Illustration of the rules of the Chinese CS system and the proposed framework (UniCUE). (a) The chart for Mandarin Chinese CS (figure from \protect\cite{liu2019pilot}), where five distinct hand positions are used to encode vowels, and eight finger shapes are employed to represent consonants in Mandarin Chinese. (b) Our framework enables the direct generation of synchronized natural speech from video.} 
    \label{fig:overview}
    \vspace{-0.3cm}
\end{figure}

\section{Introduction}
Cued Speech (CS) is an visual phonetic encoding system that utilizes specific hand shapes and positions to enhance lip reading, providing an accurate visual representation of all phonemes in spoken language \cite{cornett1967cued,liu2019pilot,leybaert2010cued}. 
CS maintains a high level of consistency with spoken language in terms of phonemes and speech patterns, enabling hearing-impaired individuals to better integrate into speech-dominant social and educational environments \cite{cornett1967cued,leybaert2010cued,leybaert201019}. 
In Mandarin Chinese, CS employs 8 hand shapes and 5 positions to encode consonants and vowels (as illustrated in Figure \ref{fig:overview}(a)), addressing challenges such as the phonemes with similar lip shapes \cite{liu2019pilot}.

\textbf{CS V}ideo-to-\textbf{S}peech generation (\textbf{CSV2S}) task aims to convert CS videos of into comprehensible speech signals.
However, directly constructing an end-to-end CSV2S model faces several challenges. 
\textbf{Firstly}, this task involves complex multimodal semantic correlations, requiring precise mapping from visual cues (lip movements and hand coding) to acoustic speech, while the limited scale of existing CS datasets further constrains model capacity.
\textbf{Secondly}, fine-grained spatiotemporal modeling of visual information is essential to resolve the intrinsic asynchrony, \textit{i.e.,} the hand-preceding phenomenon, where hand cues precede corresponding lip movements~\cite{liu2020re}.
\textbf{To the best of our knowledge, the CSV2S task has not been explicitly studied in prior literature.}

Existing research primarily focuses on \textbf{CS R}ecognition (\textbf{CSR}) that converts CS videos into phoneme-level text \cite{liu2020re,liu2024computation,liu2018visual,liu2019automatic}, neglecting the critical need for natural speech generation. 
This limitation significantly impairs real-time communication between hearing-impaired and normal-hearing individuals, especially in educational and social scenarios.
For instance, in group conversations, normal-hearing participants must quickly comprehend and respond to questions posed by their hearing-impaired peers.
Textual output from CSR systems is often insufficient for such natural and smooth interactions.
Additionally, recent lipreading-based video-to-speech models such as LipVoicer~\cite{yeminilipvoicer} rely solely on lip movements, failing to capture the complementary hand-coded information in CS that conveys critical phonemic distinctions.
These shortcomings underscore the need for a more comprehensive approach.
Motivated by this, we aim \textbf{to develop the first Chinese CSV2S system that directly decodes CS videos into intelligible speech}, as illustrated in Figure~\ref{fig:overview}(b).

\begin{figure}[t]
    \centering
    \includegraphics[width=0.99\linewidth]{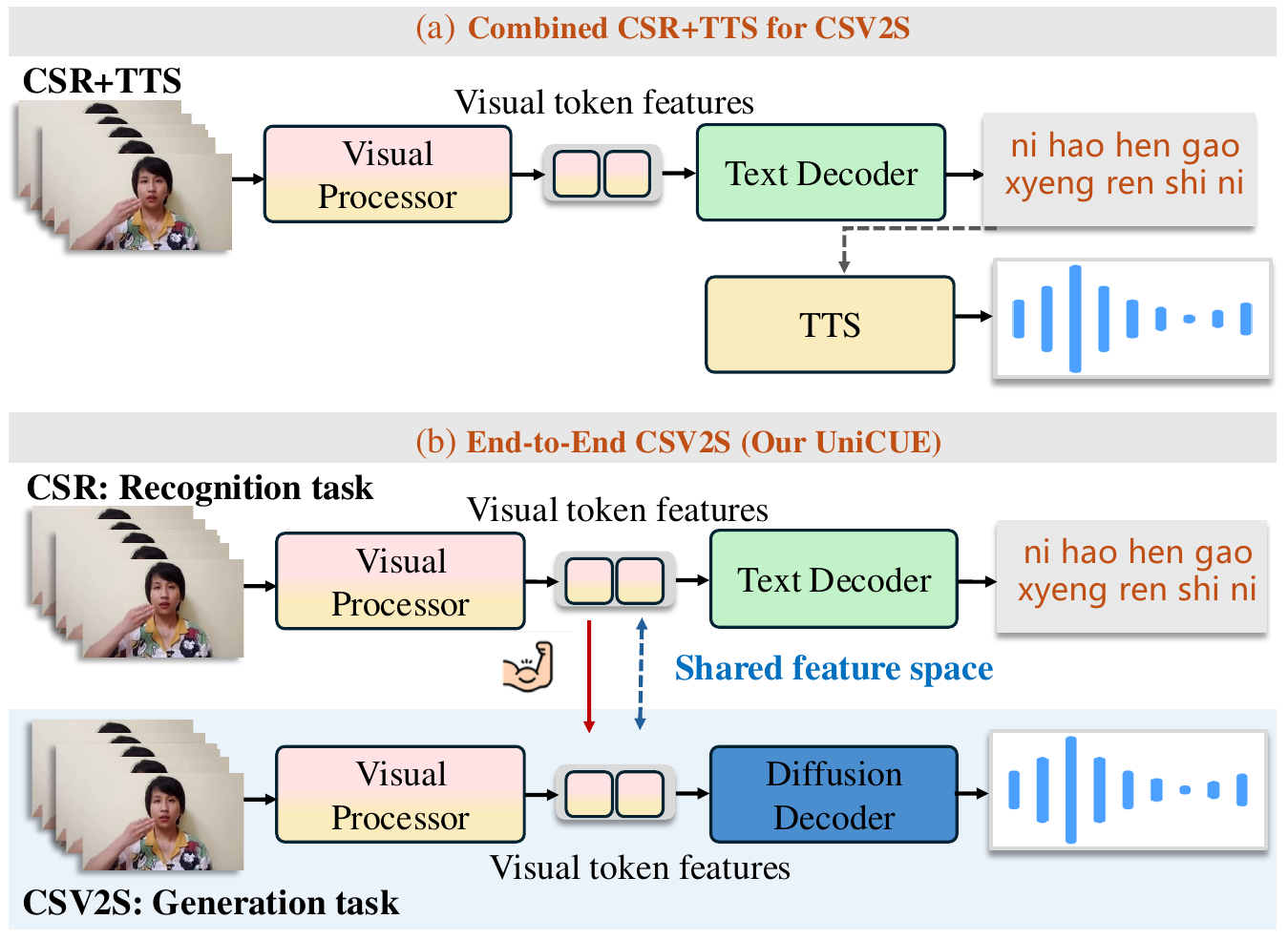}
    \caption{(a) The combined CSV2S architecture combines separately trained CSR and TTS models. (b) Our unified framework (UniCUE) that transfers understanding capabilities of CSR into speech generation training by integrating the visual processor of CSR into CSV2S. }
    \label{fig:motivation}
\end{figure}

 A straightforward solution, shown in Figure~\ref{fig:motivation}(a), is to combine a CSR model with a Text-to-Speech (TTS) system. However, this combined pipeline suffers from two key drawbacks. 
 Firstly,  the intermediate textual representation introduces error propagation, as misrecognitions in the CSR stage lead to incorrect speech output.
Secondly, the textual intermediate discards fine-grained spatiotemporal cues in the CS video, resulting in synthesized speech that lacks temporal alignment with the visual input. 

\textbf{To overcome these challenges}, we draw inspiration from recent advances in multimodal learning, where semantic reasoning from vision-language models (VLMs) has shown strong promise in tasks like
text-guided image synthesis with interleaved control \cite{mi2025think,chen2025multimodal}.
We hypothesize that the multimodal visual understanding inherent in CSR can serve as a semantic bridge to support more accurate and controllable speech generation in CSV2S.
As depicted in Figure~\ref{fig:motivation}(b), we introduce a unified framework that leverages a shared visual processor to bridge CSR (understanding task) and CSV2S (generation task). This processor serves as a two-way translator: during CSR, it extracts linguistic semantics from fine-grained lip-hand motion patterns; in CSV2S, it utilizes these semantics to guide speech generation.
The core innovation of our framework lies in modeling a semantic compensation flow, where phoneme-level supervision from CSR reduces ambiguity in speech synthesis, enabling more faithful and coherent voice generation under complex multimodal conditions.

 Building upon this semantic compensation paradigm, in this work, we propose \textbf{UniCUE}, the first unified framework that bridges CSR and CSV2S tasks through three specific components:
\textbf{Firstly}, unlike prior CSR methods \cite{liu2024computation,liu2023cross} that process lip and hand modalities independently and rely on raw video embeddings, UniCUE employs a pose-aware processor that fuses video and pose streams into a mixed representation. This enables fine-grained spatiotemporal modeling of the hand-preceding phenomenon and improves generalization to cuer-specific expressive styles.
\textbf{Secondly}, to enhance the alignment between visual and linguistic semantics, we introduce a semantic alignment pool to map the video and pose latent spaces into a shared textual space using contrastive learning. This facilitates cross-modal correlation modeling and improves semantic consistency in the generated speech.
\textbf{Thirdly}, 
to unify the understanding and generation tasks, we reuse the CSR visual encoder within our diffusion-based CSV2S decoder and introduce a VisioPhonetic Adapter (VPA) that transforms the visual representations into diffusion-compatible codes. This design enables the decoder to effectively incorporate fine-grained semantic information derived from multimodal visual inputs

To evaluate UniCUE on hearing-impaired individuals, we extend the MCCS dataset \cite{lei2024bridge} by adding data from 8 hearing-impaired and 2 normal-hearing cuers\footnote{Cuer means the people who perform CS.}, forming the Unified-HI Corpus with 14 cuers.

Experimental results on this dataset demonstrate that UniCUE not only produces accurate and intelligible speech, but also maintains temporal synchronization with the CS video.

The main contributions of this work can be summarized as:
\begin{itemize}
    \item We propose the first CSV2S framework by constructing a unified multimodal system that integrates CSR capabilities to enhance speech generation.
    \item We propose a pose-aware visual processor and a semantic alignment pool to enhance fine-grained, semantically aligned visual representations, and introduce an VPA module to convert fine-grained semantic information into understandable coding for the speech synthesis model.
    \item We construct a new Mandarin Chinese CS dataset comprising both hearing-impaired and normal-hearing cuers. Experimental results demonstrate that our UniCUE outperforms the state-of-the-art (SOTA) methods in terms of speech accuracy, consistency, and quality.
\end{itemize}

\begin{figure*}[t]
    \centering
    \includegraphics[width=0.9\textwidth]{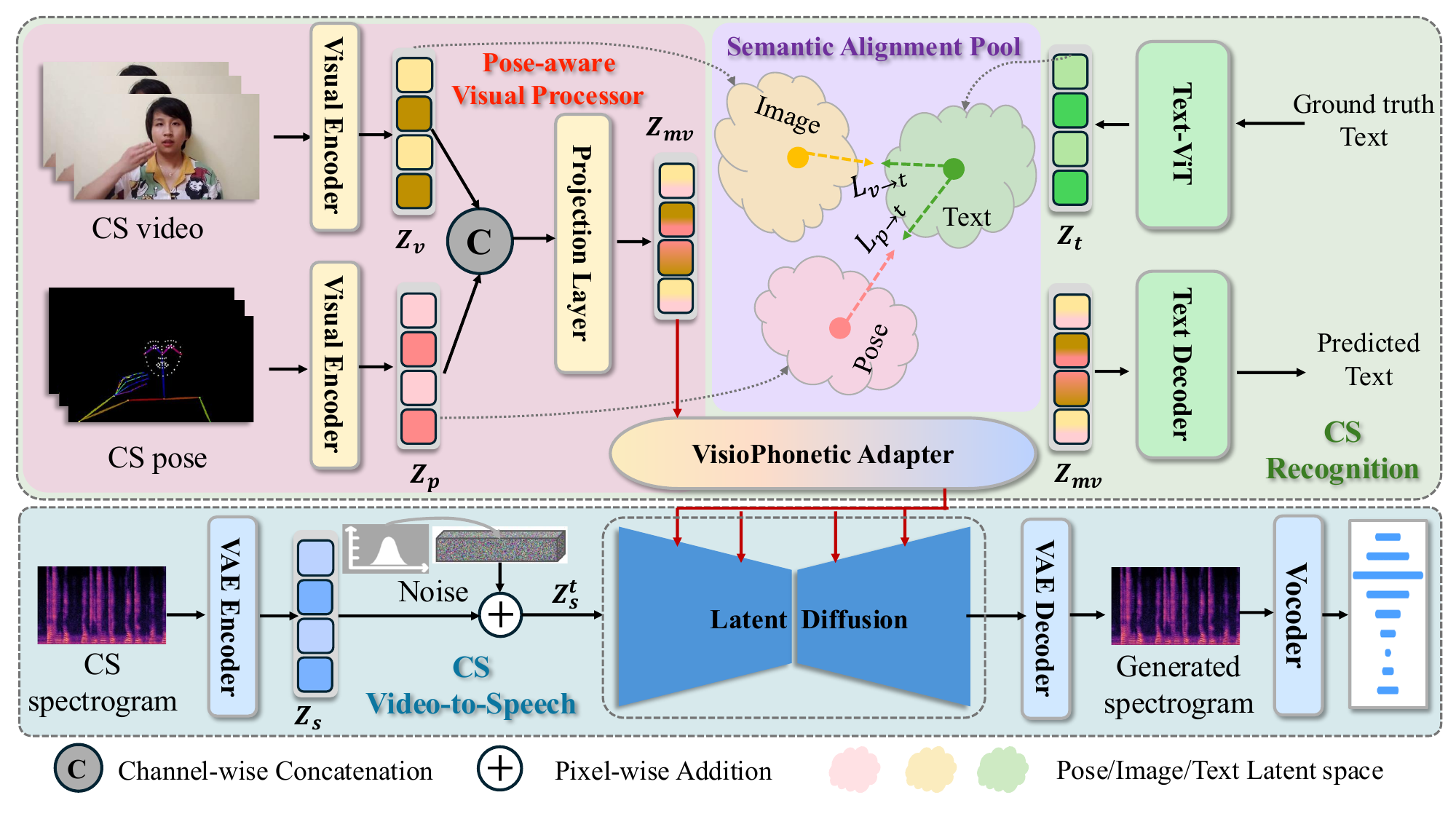}
    \caption{\textbf{Overview of our unified framework (UniCUE).} It achieves direct Chinese CSV2S generation with
    semantic consistency, temporal alignment, and characteristics coherence by aligning the fine-grained spatiotemporal visual representations of CSR with the diffusion-based speech generator. The framework consists of three core modules:
    (1) \textcolor{myred}{Pose-Aware Visual Processor}: Integrates video and pose embeddings to perform fine-grained spatiotemporal modeling of lip and hand movements.
    (2) \textcolor{mypurple}{Semantic Alignment Pool}: Enhances the semantic mapping between visual features and speech content through video-text and pose-text contrastive learning.
    (3) \textcolor{myblue}{VisioPhonetic Adapter (VPA)}: Converts fine-grained visual representation of CSR into condition encodings  compatible with the diffusion-based generator.
    }
    \label{fig:framework}
\end{figure*}

\section{Related Work}

\subsection{Video-to-Speech Generation}
V2S aims to synthesize natural speech aligned with silent talking videos, but is challenged by limited data. Uni-Dubbing~\cite{lei2024uni} addresses this via modality-aligned pre-training on multimodal data and fine-tuning with both multimodal and audio-only inputs. Similarly, Kefalas \textit{et al.}~\cite{kefalas2024large} pre-train on large audio-only corpora before tuning on paired data.
Some studies~\cite{kim2023lip,yeminilipvoicer,gupta2024visatronic} incorporate transcripts to enhance generation. Kim \textit{et al.}~\cite{kim2023lip} use text-speech supervision to improve word-level representation via multi-task learning.
Existing V2S methods primarily focus on lip reading. However, CS conveys phonemic information through both lip and hand movements. Ignoring hand cues results in incomplete visual representations and degraded speech synthesis quality, limiting the applicability of these methods to CS. Notably, no prior work has addressed the CSV2S task.

\subsection{Cued Speech Recognition}
CS augments lip reading with hand coding to support the hearing-impaired. The CSR task aims to transcribe CS videos into text by leveraging lips and hands as complementary modalities~\cite{liu2019pilot,papadimitriou2021fully}. 
Most CSR methods extract lip and hand features separately and fuse them for recognition~\cite{liu2020re,liu2024computation,liu2023cross,zhang2023cuing}. Due to the asynchronous nature of these modalities, effective fusion remains challenging. Liu \textit{et al.}~\cite{liu2020re} proposed re-synchronization to align hand with lip features, while transformer-based mutual learning~\cite{liu2023cross,liu2024computation} improves multimodal interaction. Zhang \textit{et al.}~\cite{zhang2023cuing} addressed privacy concerns via federated learning.
In contrast, we directly model lip and hand cues from whole frames, avoiding explicit fusion. A pose-aware visual processor is introduced further to enhance cross-modal representation and improve performance.

\subsection{Unified Understanding and Generation}
Recent advances in unifying understanding and generation tasks fall into two main paradigms. The first integrates visual-language understanding with external generative models (\textit{e.g.,} diffusion models) for multimodal generation \cite{wu2024next,dong2024dreamllm,jin2024unified,li2024mini,sun2024generative,ge2024seed,ge2025divot}. For example, \cite{jin2024unified,li2024mini} utilize large language models (LLMs) for semantic understanding and diffusion models \cite{rombach2022high,podell2023sdxl} for high-fidelity image synthesis.
The second paradigm trains LLM-based foundation models via next-token prediction for both vision understanding and generation \cite{yu2023scaling,sun2023emu,zhoutransfusion,wu2024vila,fang2024puma,wu2025janus,chen2025janus}. Transfusion \cite{zhoutransfusion}, for instance, unifies image understanding and generation within a single transformer, enabling controllable text-to-image synthesis by preserving visual details.
However, existing approaches mainly focus on visual-text settings, leaving visual-to-speech generation underexplored. In this work, we introduce the first unified framework that bridges visual understanding and speech generation.

\section{Method}

\subsection{Overview of UniCUE}
 \label{subsec:unified}
 To achieve accurate CSV2S generation, the proposed method needs to simultaneously address two critical challenges: (1) \textbf{semantic understanding} of the linguistic correlations between visual cues and speech content, and (2) \textbf{speech synthesis} that preserves cuer-specific characteristics and temporal alignment. 
Inspired by the auxiliary benefits of unified understanding and generation for multi-modal controllable image synthesis \cite{wu2024next,dong2024dreamllm}, we design a unified architecture that integrates CSR and CSV2S, enabling CSV2S with understanding capability improvement through shared visual feature representations.
As illustrated in Figure \ref{fig:framework}, the framework operates via two pathways.

\noindent\textbf{CSR: Fine-grained Visual Cues Understanding.}  
As the recognition pathway, CSR models fine-grained spatiotemporal visual semantics to transcribe CS videos into linguistic sequences.  
Given a CS video \( I_v \) and its corresponding pose maps \( I_p \) (extracted via OpenPose~\cite{8765346}), we first utilize a pose-aware visual processor to extract multi-modal embeddings \( Z_{mv} \), which capture lip and hand motion cues. And then \( Z_{mv} \) is fed into a auto-regressive Transformer-based text decoder \( D_T \), which models long-range dependencies and contextual interactions across the sequence to generate the predicted token sequence: \( T_p = D_T(Z_{mv}) \),  
where \( T_p \) denotes the predicted token sequence.  

Unlike prior approaches relying on 
Connectionist Temporal Classification (CTC) loss~\cite{graves2006ctc}, which predict each token independently and thus limit the model’s ability to capture cross-token dependencies and coarticulatory effects, our method employs an auto-regressive decoder \( D_T \) supervised by cross-entropy loss.
This design allows \( D_T \) to generate tokens  conditioned on previously generated outputs and spatialtemporal visual cues, which is more suited to modeling the asynchronous and dynamic nature of CS.

To further enhance both token-level precision and sequence-level linguistic consistency,
we employ a hybrid training objective:
a masked language modeling loss \( \mathcal{L}_{\text{CE}}^{\text{masked}} \) supervises selectively masked ground-truth tokens to enhance contextual understanding; a sequence-level cross-entropy loss \( \mathcal{L}_{\text{CE}}^{\text{seq}} \)enforces supervision over the full sequence to promote accurate transcription.
The final training objective for CSR is:
\begin{equation}
\mathcal{L}_{R} = \mathcal{L}_{\text{CE}}^{\text{masked}}(T_p, T_g) + \mathcal{L}_{\text{CE}}^{\text{seq}}(T_p, T_g),
\label{eq:csr-loss}
\end{equation}
where \( T_g \) denotes the ground-truth token sequence.
This dual-loss strategy enhances token-level accuracy while preserving global sequence semantics, enabling the model to capture subtle visual-linguistic cues and temporal dynamics inherent in CS videos, thus improving recognition performance and supporting speech synthesis.

\noindent\textbf{CSV2S: Cuer-specific Speech Synthesis.}  
To directly synthesize intelligible and personalized speech from CS videos, we formulate speech generation as a conditional denoising process within a latent diffusion model (LDM)~\cite{rombach2022high}.  
Since both lip shapes and hand cues in CS convey phonemic content, the speech generation is conditioned a refined visual embedding \( Z_{mv}' \), which is derived by transforming the CSR multimodal feature \( Z_{mv} \) via a VisioPhonetic adapter (VPA).
Specifically, a pretrained VAE encoder compresses ground-truth mel-spectrograms into latent codes \( Z_s \), 
which are progressively corrupted with Gaussian noise \( \epsilon \) over \( t \) steps:
 $Z_s^t := \alpha_t \cdot Z_s + (1 - \alpha_t) \cdot \epsilon,$
where \( \alpha_t \) denotes the noise level at timestep \( t \).  
The noisy latent \( Z_s^t \) then denoised by the LDM conditioned on \( Z_{mv}' \). The generation objective is defined as:
\begin{equation}
\mathcal{L}_{G} := \mathbb{E}_{Z_s^t, \, Z_{mv}, \, \epsilon, \, t} \left[ \left\| \epsilon - \mathcal{M}(Z_s^t, Z_{mv}, t) \right\|^2_2 \right],
\label{eq:V2S}
\end{equation}
where \( \mathcal{M} \) represents the denoising network.  
By learning this conditional distribution, our model generates temporally aligned speech that reflects the visual expressions of cuers.  

\noindent\textbf{UniCUE: Unified Understanding and Generation.}
The CSR pathway learns fine-grained multi-modal visual embeddings \( Z_{mv} \) through detailed linguistic recognition. To bridge the architectural gap between the CSR and the diffusion-based speech generator, we introduce a VPA that transforms \( Z_{mv} \) into a refined representation \( Z_{mv}' \).  
These embeddings are subsequently utilized as conditional inputs to the CSV2S pathway, enabling the speech synthesis model to leverage enriched visual understanding for improved generation accuracy. By sharing visual feature representations within this unified framework, our approach effectively reduces information loss and mitigates error propagation that often arises from intermediate text conversions. 
As a result, CSV2S is capable of generating cue-specific speech that faithfully preserves linguistic fidelity and temporal alignment, producing personalized and intelligible speech outputs tailored to individual cuers.

\subsection{Pose-aware Visual Processor}
\label{subsec:visual processor}

Considering the strong spatiotemporal correlation between hand coding, lip movement, and their underlying semantic content, both CSV2S and CSR require accurate modeling of lip and hand motion patterns. This necessitates a visual encoder capable of capturing fine-grained and temporally coherent features. 
While video frames offer rich appearance information, they often suffer from redundancy and visual ambiguity. In contrast, pose maps provide a compact, structured, and noise-resilient representation of motion dynamics.
To leverage the complementary strengths of both modalities, we design a pose-aware visual processor that constructs fused visual representations, as shown in Figure~\ref{fig:framework}.

Specifically, the input to the processor consists of video frames \( I_v \) and pose maps \( I_p \), both formatted as tensors of shape \( T \times 3 \times H \times W \), where \( T \) indicates the frame lengths, and \( H \times W \) denotes the spatial resolution.
The processor comprises two main components. First, a shared visual encoder \( E_V \) extracts spatiotemporal features from both modalities via a sequential architecture: a 2D ResNet backbone extracts frame-wise spatial features, which are stacked along the temporal axis and passed through a 1D temporal convolution to model short-term motion patterns. The resulting sequence is then fed into a Transformer encoder to capture long-range temporal dependencies across frames. This process yields the video features \( Z_v = E_V(I_v) \) and pose features \( Z_p = E_V(I_p) \), where
$Z_v \in \mathbb{R}^{L \times D}$, $Z_p \in \mathbb{R}^{L \times D}$
with \( D \) denoting the embedding dimension and \( L = T \times N \) being the total number of tokens, where \( N \) is the number of spatial patches per frame.
Second, the projection layer integrates the two feature streams. The video and pose features are concatenated along the channel dimension and passed through a multi-layer perceptron (MLP), consisting of two linear layers with ReLU activation and LayerNorm, to produce the final mixed visual representation:
\begin{equation}
    Z_{mv} = \text{MLP}(\text{Concat}(Z_v, Z_p)).
\end{equation}
The fused representation \( Z_{mv} \) serves as a unified visual embedding that drives both recognition and generation pathways. In the subsequent modules, this representation is semantically aligned with linguistic content and refined for diffusion-based speech synthesis.

\subsection{Semantic Alignment Pool}
\label{subsec:alignment}

To further enhance semantic consistency between visual representation and linguistic content, we introduce a semantic alignment mechanism that aligns video, pose, and textual modalities through contrastive learning.
Specifically, a ViT-based text encoder encodes the ground-truth transcript tokens \( T_g \) into text embeddings \( Z_t \). The visual features \( Z_v \) and pose features \( Z_p \), extracted by the pose-aware visual processor, are projected into a shared latent space via learnable linear layers. The text embedding \( Z_t \) is similarly projected.
We adopt a contrastive loss across the batch, treating each video-text and pose-text pair from the same sample as a positive pair, and all others as negatives. The loss is denoted as:
\begin{equation}
\mathcal{L}_{v \leftrightarrow t} = 1 - \cos(Z_v, Z_t), \quad
\mathcal{L}_{p \leftrightarrow t} = 1 - \cos(Z_p, Z_t),
\end{equation}
where \( \cos(\cdot,\cdot) \) denotes the cosine similarity between normalized embeddings.
The total semantic alignment loss is calculated as:
\begin{equation}
    \mathcal{L}_{S} = \mathcal{L}_{v \leftrightarrow t} + \mathcal{L}_{p \leftrightarrow t}.
    \label{eq:CL}
\end{equation}
By enforcing this high-level alignment, the model is encouraged to extract complementary and discriminative semantics from visual modalities.These aligned features not only enhance linguistic recognition in CSR, but also offer semantically grounded condition for accurate speech synthesis.

\begin{figure}[t]
    \centering
    \includegraphics[width=0.5\linewidth]{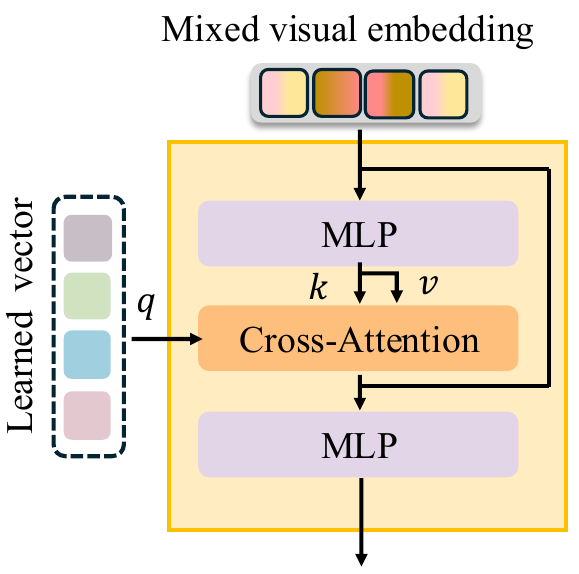}
    \caption{The details of the VisioPhonetic Adapter, which transforms semantic visual embeddings into phonetic-aware features to enable seamless conditioning for diffusion-based speech synthesis.}
    \label{fig:VPA}
\end{figure}

\subsection{VisioPhonetic Adapter}
While the CSR-derived embeddings capture rich visual-linguistic semantics, they remain mismatched in format and granularity for direct use in diffusion-based speech generation. To bridge this modality gap, we propose the  \textbf{V}isio\textbf{P}honetic \textbf{A}dapter (VPA), which transforms semantically aligned visual features into a phonetic-aware conditioning signal suitable for the LDM.
As illustrated in Figure \ref{fig:VPA}, this lightweight module employs a sequential architecture to progressively refine visual-semantic representations into a diffusion-compatible conditioning signal:
\begin{equation}
\mathbf{Z}_{mv}^{\prime} = \text{MLP}\Big(\text{CrossAttn}\big(\text{MLP}(\mathbf{Z}_{mv})\big)\Big),
\end{equation}
which includes two MLPs and a Q-Former-style~\cite{li2023blip} cross-attention layer.
We use \( N_q \) learnable semantic queries \( \mathbf{f} \in \mathbb{R}^{N_q \times D} \), which is initialized by computing the average latent representation from ground-truth mel-spectrograms encoded by the pretrained VAE.
 This provides a phonetic-aware initialization aligned with the diffusion model’s target space.
These queries act as phonetic slots to extract and reorganize relevant patterns from \( \mathbf{Z}_{mv} \). The cross-attention mechanism operates as: 
$\mathbf{q} = \mathbf{W}^q \mathbf{f}, 
\mathbf{k} = \mathbf{W}^k \mathbf{Z_{mv}}, 
\mathbf{v} = \mathbf{W}^v \mathbf{Z_{mv}}, 
\mathbf{a} = \text{Softmax}\left(\frac{\mathbf{q} \mathbf{k}^T}{\sqrt{d}}\right) \mathbf{v}, \mathbf{{Z_{mv}}^{\prime}} = \text{MLP}(\mathbf{Z_{mv}}+a).$
The adapted features $\mathbf{Z}_{mv}^{\prime}$ 
serve as the final interface between visual understanding and speech synthesis, ensuring that the generated audio is not only temporally coherent but also linguistically faithful to the video input.

  \begin{table*}[t]
\centering
\caption{Comparison between our Chinese Mandarin CS dataset and existing CS dataset. \textbf{H} denotes the cuers with normal hearing, while \textbf{HI} indicates hearing-impaired cuer. Our newly proposed Unified-HI Corpus is the first large-scale Chinese CS dataset with both hearing-impaired and normal-hearing cuers.
 }
\label{tab:dataset}
\begin{tabular}{ccccccc} 
 \Xhline{1pt} 
Dataset & Cuers & Sentences &Character& Word&Resolution & FPS  \\ 
 \Xhline{1pt} %
French CS \cite{liu2018visual} & 1-H & 238 & 12872& -&$720 \times 576$ & 50 \\
British CS \cite{liu2019automatic} & 1-H & 97 & 2741&-&$720 \times 1280$ & 25 \\
MCCS \cite{lei2024bridge} & 4-H& 4000 &131608&42256& $720 \times 1280$ & 30 \\
\midrule %
 Unified-HI (Ours) & \makecell[c]{6-H and 8-HI} & 11282 &350333& 112664&$720 \times 1280$ & 30 \\
 \Xhline{1pt}  %
\end{tabular}
\end{table*}

\begin{table*}[htbp]
    \centering
    \caption{Comparison with SOTA methods on test data of normal-hearing cuers and hearing-impaired cuers.
    \textbf{Bold} and \underline{underlined} results are the best and second-best results. $\uparrow$ indicates that larger values are better, while $\downarrow$ indicates that smaller values are preferable.  }
    \begin{tabular}{l|ccccc|cccc}
        \Xhline{1pt}  
        \multirow{2}{*}{Method} & \multicolumn{5}{c|}{Normal-hearing cuers} & \multicolumn{4}{c}{Hearing-impaired cuers} \\
        \cline{2-10}
        & WER $\downarrow$& LSE-C $\uparrow$ & LSE-D$\downarrow$ & DNSMOS $\uparrow$&STOI $\uparrow$&
        WER$\downarrow$& LSE-C$\uparrow$ & LSE-D$\downarrow$& DNSMOS $\uparrow$\\
        \Xhline{1pt}  
           GT & - & 7.274 & 7.314&2.79 &- & -&-&-&-\\
        \hline
        CMML & 0.663 &4.135  &9.241& 1.24 &0.11&0.924&2.141&10.132&1.03\\
        EcoCued&0.657 &4.327 &9.146&1.28&0.12&0.917& 2.165& 10.079& 1.07\\
         \hline
         CSR (Ours)&\textbf{0.186} &4.874  & 9.125 &\textbf{2.53}&\textbf{0.57}&\textbf{0.224}& 3.342& 9.315&\textbf{2.29}\\
         \Xhline{1pt}
         Lip2Speech &0.803&4.215& 9.367&1.03&0.05&0.989&2.424&10.816&0.02\\
         LipVoicer &0.754&4.361&9.226&1.12&0.08&0.971&2.623&10.517&0.04\\
      
         \hline
         CSV2S (Ours)& 0.374&\underline{6.245}&\underline{7.962}&2.27&0.42&0.422&\underline{5.938}&\underline{8.347}&2.04\\
         UniCUE (Ours) & \underline{0.205}& \textbf{6.729}& \textbf{7.632}&\underline{2.46}&\underline{0.53}&\underline{0.248} &\textbf{6.491}&\textbf{8.076}&\underline{2.17}\\
         \Xhline{1pt}  
    \end{tabular}
    \label{tab:comparison}
\end{table*}

\section{Experiment}
\subsection{Experimental Setting}
\noindent\textbf{Dataset.} 
Existing CS datasets are limited to normal-hearing cuers and lack data from hearing-impaired individuals, hindering model generalization to the primary users of assistive communication systems. 
To bridge this gap, we construct a new dataset, the \textbf{Unified-HI Corpus}, which includes CS videos from 8 hearing-impaired and 6 normal-hearing cuers.
This diverse composition significantly enriches variations in gesture styles, lip movements, and speech patterns. The expanded coverage introduces more realistic challenges and better reflects practical use cases, enabling models to capture cue-specific nuances essential for hearing-impaired users. A comparison with existing CS datasets is shown in Table~\ref{tab:dataset}, 
and further details on sentence coverage and phoneme distribution are included in \textbf{Appendix Section 2}.

Due to the noisy speech data from hearing-impaired cuers, we use CS data from 6 normal-hearing cuers for training. The data from normal-hearing cuers is split by sentence into training and test sets with a 95:5 ratio to ensure effective training and validation. Importantly, \textbf{all CS data from the 8 hearing-impaired cuers are used in the test set}, enabling a robust evaluation of model generalization to this group.

\noindent\textbf{Architecture Details.}  
The CSV2S pathway is entirely built upon the AudioLDM ~\cite{liu2023audioldm}, including its VAE encoder-decoder, latent diffusion model, and vocoder components.
For CSR, the Transformer in visual process, tokenizer, text-ViT, and text decoder are initialized from MBart \cite{liu2020multilingual}.  
Detailed training and inference configurations are provided in \textbf{Appendix Section 1}.

\noindent\textbf{Evaluation Metrics.}  
We evaluate the synthesized speech from three perspectives: linguistic accuracy, temporal synchronization, and speech quality.
Linguistic accuracy is quantified by the Word Error Rate (WER) between the recognized text and ground truth.
Temporal synchronization is assessed using SyncNet~\cite{Chung16a}, reporting LSE-D (temporal distance) and LSE-C (confidence score).
Speech quality is evaluated via STOI~\cite{taal2010short} for intelligibility and DNSMOS~\cite{reddy2021dnsmos} for naturalness.

\noindent\textbf{Comparison Methods.}  
We evaluate our \textbf{UniCUE} against:  
(1) \textbf{CSV2S (Ours)}: direct speech synthesis without CSR assistance;  
(2) \textbf{CSR (Ours)}: including pose-aware visual processor, text encoder and decoder, and semantic alignment pool;  
(3) \textbf{CSR methods}: CMML~\cite{liu2023cross} and EcoCued~\cite{liu2024computation};  
(4) \textbf{V2S methods}: Lip2Speech~\cite{choi2023intelligible} and LipVoicer~\cite{yeminilipvoicer}.

\subsection{Comparison with SOTA Methods}

\noindent\textbf{Quantitative Comparison.}  
We compare our framework against SOTA methods, as summarized in Table \ref{tab:comparison}. Our CSR model, empowered by the pose-aware visual processor and semantic alignment pool, achieves significantly lower WERs (0.186 for normal-hearing and 0.224 for hearing-impaired cuers), surpassing previous CSR methods. Building on this strong semantic understanding, UniCUE 
outperforms V2S methods across LSE-D, LSE-C, DNSMOS, and STOI metrics, demonstrating superior linguistic accuracy, temporal alignment, and speech quality. 

\noindent\textbf{Qualitative Comparison.}  
Mel-spectrogram visualizations (\textbf{Figure 4 in Appendix}) further highlight the advantages of our method, showcasing improved temporal synchronization and clearer acoustic structures compared to others.

\begin{table*}[htbp]
    \centering
    \caption{Ablation Studies of model components on test data of norma hearing cuers and hearing-impaired cuers. The notations X\textsuperscript{††}, X\textsuperscript{‡}, and X\textsuperscript{*} 
    indicate ablated versions of the architecture X, where the pose maps, semantic alignment pool, and VPA module are removed, respectively.}
    \begin{tabular}{l|ccccc|cccc}
        \Xhline{1pt}  
        \multirow{2}{*}{Method} & \multicolumn{5}{c|}{Normal-hearing cuers} & \multicolumn{4}{c}{Hearing-impaired cuers} \\
        \cline{2-10}
        & WER $\downarrow$& LSE-C $\uparrow$ & LSE-D$\downarrow$ & DNSMOS $\uparrow$&STOI $\uparrow$&
        WER$\downarrow$& LSE-C$\uparrow$ & LSE-D$\downarrow$& DNSMOS $\uparrow$\\
        \Xhline{1pt}  
           GT & - & 7.274 & 7.314&2.79 &- & -&-&-&-\\
        \hline CSR\textrm{\textsuperscript{†}}\textrm{\textsuperscript{†}}&0.210 & 4.746&9.129 & 2.42&0.49&0.250&3.218&9.402&2.19\\
        
         CSR\textrm{\textsuperscript{‡}}&0.204 & 4.783&9.224 &2.46 &0.53&0.247&3.234 &9.397&2.21\\
         
         CSR &\textbf{0.186} &4.874  & 9.125 &\textbf{2.53}&\textbf{0.57}&\textbf{0.224}& 3.342& 9.315&\textbf{2.29}\\
         \hline
         CSV2S\textrm{\textsuperscript{†}} &0.398 &6.158& 8.122&2.21& 0.40&0.398&5.821&8.582&1.96  \\
        CSV2S&0.374&{6.245}&{7.962}&2.27&0.42&0.422&{5.938}&{8.347}&2.04 \\
         \hline
         UniCUE\textrm{\textsuperscript{†}}\textrm{\textsuperscript{†}}& 0.239& 6.637& 7.724& 2.30&0.44&0.267&6.419&8.163&2.08\\
        
        UniCUE\textrm{\textsuperscript{‡}}& 0.231&\underline{6.641}&\underline{7.716}&2.33&0.46&0.276&\underline{6.421}&\underline{8.159}&2.10\\
        
        UniCUE\textsuperscript{*} &0.226 &6.613&7.731&2.37&0.48&0.271&6.410& 8.167&2.12 \\ 
        
         UniCUE & \underline{0.205}& \textbf{6.729}& \textbf{7.632}&\underline{2.46}&\underline{0.53}&\underline{0.248} &\textbf{6.491}&\textbf{8.076}&\underline{2.17}\\
         \Xhline{1pt}  
    \end{tabular}
    \label{tab:ablation}
\end{table*}

\begin{figure*}[t]
    \centering
    \includegraphics[width=0.85\linewidth]{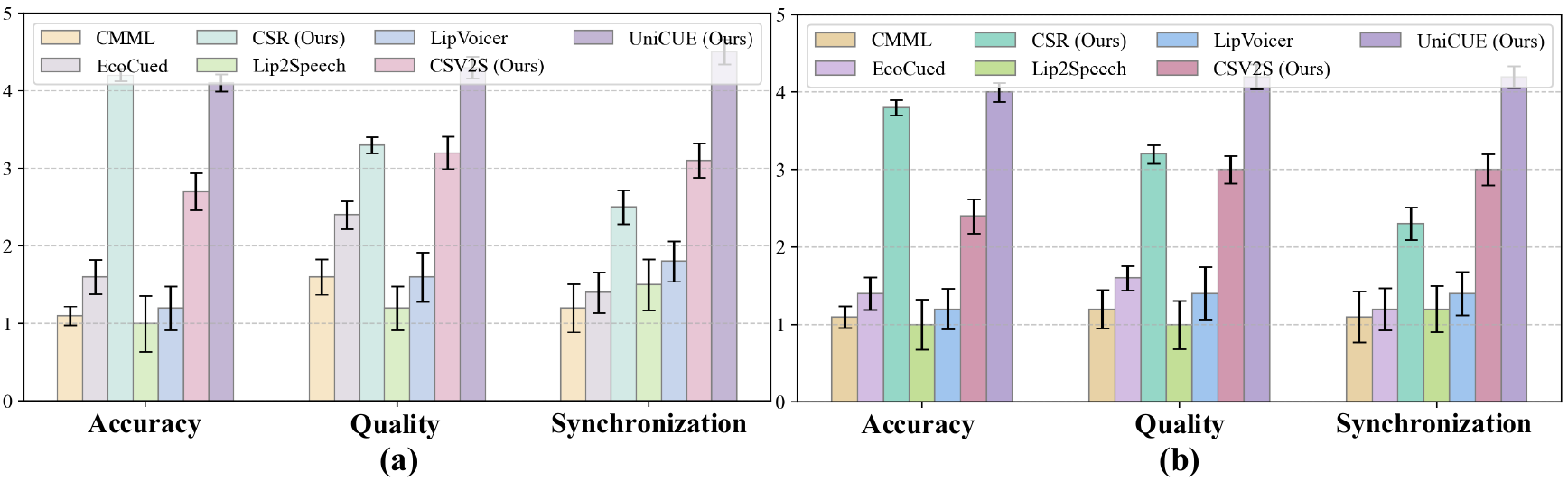}
    \caption{\textbf{User study} results for accuracy, quality, and synchronization metrics on normal-hearing (a) and hearing-impaired (b) test samples.}
    \label{fig:user_study}
\end{figure*}

\subsection{Ablation Studies}
To verify the contribution of each component, we conduct ablation studies on both normal-hearing and hearing-impaired test data. Results are summarized in Table \ref{tab:ablation}.

\noindent\textbf{Unified Training Paradigm.}
Compared to direct CSV2S, UniCUE reduces WER by 45\% (0.205 vs. 0.374) on normal-hearing cuers and 41\% (0.248 vs. 0.422) on hearing-impaired cuers. 
These results highlight the benefit of leveraging fine-grained visual semantics from CSR to enhance CSV2S, alleviating the challenge of modeling complex multimodal correlations.

\noindent\textbf{Visual Processor Design.}  
Models that rely solely on raw video features struggle to capture fine-grained motion due to redundant and noisy visual information, resulting in suboptimal performance.  
By incorporating pose cues, our visual processor effectively captures cuer-specific dynamics, leading to significantly improved accuracy and robustness across diverse cuers. 
\noindent\textbf{Semantic Alignment Mechanism.}
Disabling the Semantic Alignment Pool (SAP) degrades visual-semantic consistency, resulting in higher WERs for both CSR and UniCUE. This underscores the importance of the alignment in enforcing spatiotemporal coherence between visual cues and phonemic representations for accurate semantic modeling. The effectiveness of SAP is further validated by the t-SNE visualizations (\textbf{Figure 5 in Appendix}).

\noindent\textbf{VisioPhonetic Adapter.}
Removing the VPA results in noticeable degradation in temporal alignment, demonstrating its crucial role in bridging the representation gap between CSR and CSV2S. By adaptively selecting and refining fine-grained spatialtemporal visual cues through learnable queries, the VPA enables more accurate and temporally coherent speech synthesis.

\noindent\textbf{Impact of Hand Cues.}
Removing hand cues leads to substantial performance degradation, particularly for hearing-impaired users who often exhibit limited oral articulation and atypical lip shapes (\textbf{see Appendix Table 1}). The results highlight the complementary role of hand gestures in enhancing visual phonemic representations for CS.

\noindent\textbf{Computational Efficiency.}
UniCUE achieves faster training convergence and 40\% lower inference time than the combined pipeline (\textbf{see details in Appendix Section 5}).

\subsection{User Study}

To comprehensively assess the perceptual quality of synthesized speech, we conduct a user study involving 20 randomly selected test samples per cuer. Twenty volunteers rate the generated speech on three perceptual dimensions using 5-point Likert scales: \textbf{Accuracy} (1: unintelligible, 5: perfectly intelligible), \textbf{Quality} (1: artificial, 5: human-like), and \textbf{Synchronization} (1: desynchronized, 5: perfectly aligned).
As shown in Figure \ref{fig:user_study}, UniCUE consistently achieves significantly higher scores across all metrics, demonstrating statistically meaningful improvements. These findings validate that our unified framework effectively bridges visual understanding and speech generation, delivering superior performance in human perception compared to both modular pipelines and task-specific baselines.

\section{Conclusion}

This work introduces UniCUE, the first unified framework for directly generating speech from CS videos. By integrating fine-grained visual understanding with diffusion-based speech synthesis, UniCUE produces intelligible speech with precise temporal alignment.
Key components including the pose-aware visual processor, semantic alignment pool, and VisioPhonetic Adapter, enable effective knowledge transfer from CS recognition (CSR) to CS video-to-speech generation (CSV2S), enhancing both linguistic accuracy and temporal synchronization.
Additionally, we introduce the UniCUE-HI corpus, a new CS dataset featuring both normal-hearing and hearing-impaired cuers. Extensive experiments on this dataset demonstrate that UniCUE outperforms SOTA methods across multiple evaluation metrics.

\section{Acknowledgments}
This work was supported by the National Natural Science Foundation of China (No. 62471420), GuangDong Basic and Applied Basic Research Foundation (2025A1515012296), and 2025 Tencent AI Lab Rhino-Bird Program.

\bibliography{aaai2026}

\section*{Appendix}
\addcontentsline{toc}{section}{Appendix}


 \begin{figure}[t]
     \centering
     \includegraphics[width=0.7\linewidth]{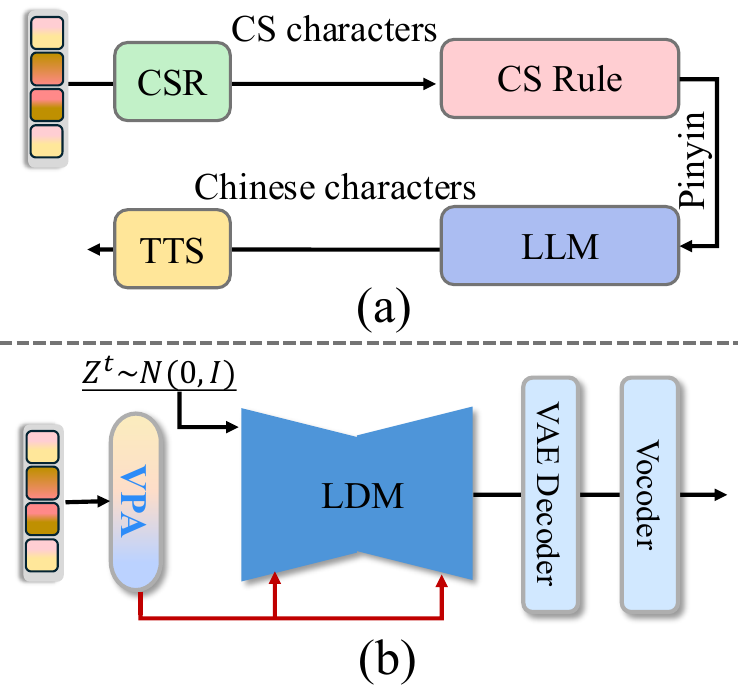}
     \caption{Comparison of CSV2S generation at inference stage. (a) The combined CS speech generation pipeline of CSR and TTS. (b) Our UniCUE. During inference, the extracted visual mixture embedding serves as a condition to guide direct speech generation.}
     \label{fig:infer}
 \end{figure}
 
\subsection{1. Details of Training and Inference}
\noindent\textbf{Training Details.} The training process is divided into two sequential phases to ensure effective knowledge transfer from visual understanding to speech synthesis. 
\noindent\textbf{Stage 1: CSR} focuses on learning fine-grained spatiotemporal representations by jointly optimizing the pose-aware visual processor, the semantic alignment pool, text encoder and decoder. This stage trains the model using the hybrid cross-entropy loss \( \mathcal{L}_{R} \) (denoted in main text Equation (1)) and the contrastive alignment loss \( \mathcal{L}_{S} \) (denoted in main text Equation (5)), formulated as: $\mathcal{L} = \mathcal{L}_{R} + \mathcal{L}_{S}.$
\noindent\textbf{Stage 2: CSV2S} leverages the learned visual representations from Stage 1 to train the VPA network and diffusion model with the  objective  \( \mathcal{L}_{G} \) (denoted in main text Equation (2)).

\noindent\textbf{Inference Details.}
During inference, the framework operates in a streamlined single-path mode to synthesize speech directly from cued CS videos. As illustrated in Figure \ref{fig:infer} (b), given an CS video and its corresponding pose, the trained pose-aware visual processor extracts the visual mixture representation \( Z_{mv} \), which encodes both semantic content and cuer characteristics. A noise sample \( Z^t \), sampled from a standard Gaussian distribution \( N(0, I) \), is fed into the trained LDM to predict the speech latent code \( \hat{Z_s} \) conditioned on the visual mixture representation over \( t \) timesteps:
\begin{equation}
  \hat{Z_s} := \frac{1}{\alpha_t} \left( Z_t - \left( 1 - \alpha_t \right) \cdot \mathcal{M}(Z^t, Z_{mv}, t) \right),  
\end{equation}
where \( \mathcal{M} \) denotes the noise prediction network of the LDM, and \( \alpha_t \) represents the scheduled variance coefficient at timestep \( t \).
Subsequently, the predicted speech latent code \( \hat{Z_s} \) is decoded into a speech mel-spectrogram via VAE decoder, and a pre-trained Vocoder \cite{liu2023audioldm} is used to convert the mel-spectrogram into a speech waveform.
Through this series of steps, the CSV2S system is able to generate natural speech that is consistent with the cuer's characteristics based on the input CS video and its corresponding pose information.

\noindent\textbf{Implementation Details.} Raw video frames and pose images are resized to \(256 \times 256\), while the audio is resampled to 16 kHz and converted into mel-spectrograms. The entire learning framework is implemented in PyTorch, and all experiments are conducted on an NVIDIA RTX A6000. We use the Adam optimizer with a learning rate of \(5 \times 10^{-4}\) for training, setting the mini-batch size to 2.

\begin{figure}
    \centering
    \includegraphics[width=0.6\linewidth]{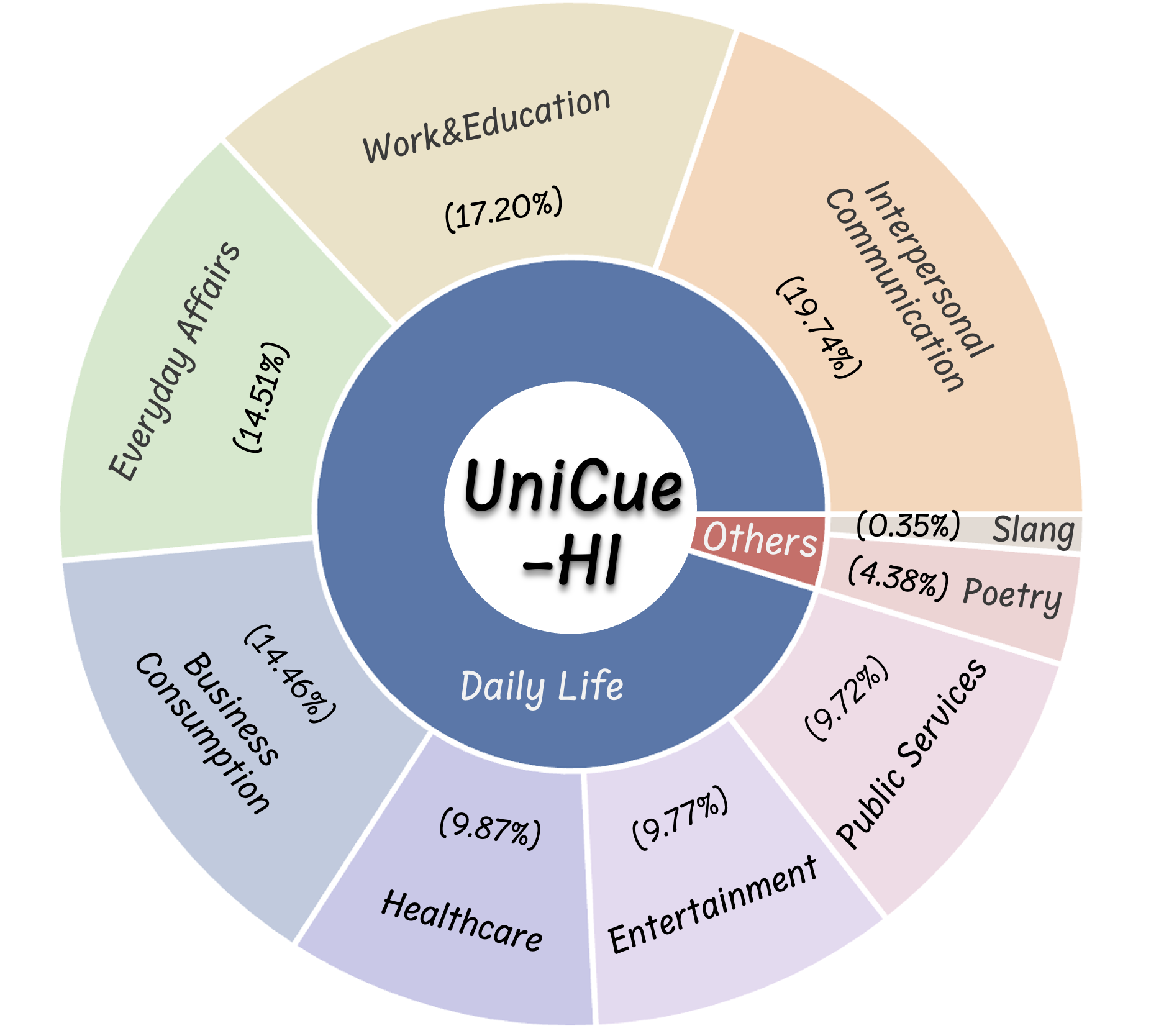}
    \caption{Statistics of sentence categories within our UniCue-HI Corpus. }
    \label{fig:dataset}
\end{figure}

\begin{figure*}
    \centering
    \includegraphics[width=0.95\linewidth]{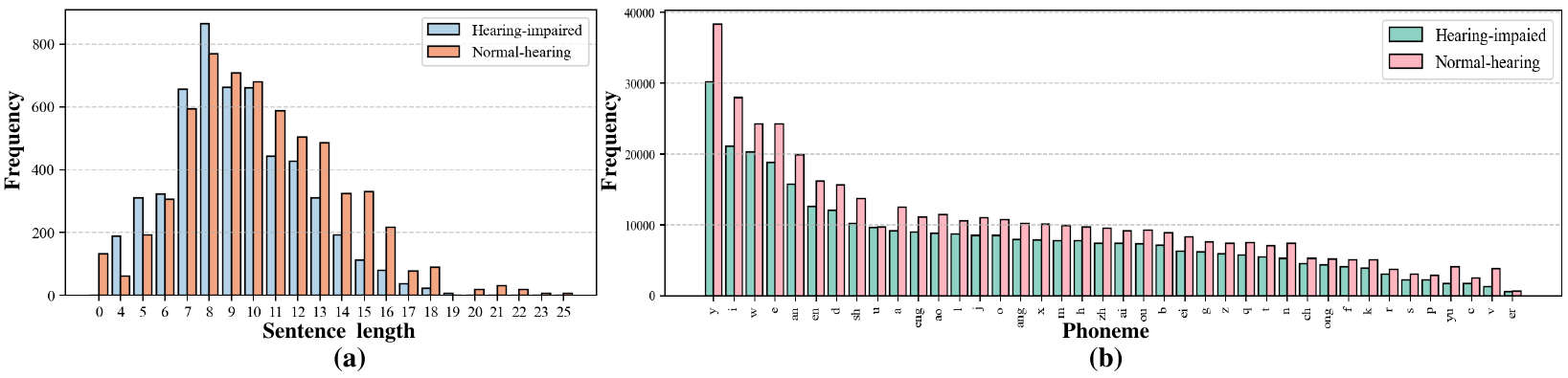}
    \caption{Statistical distributions of samples in the UniCUE-HI corpus.(a) Character-level sentence lengths; (b) Phoneme distribution. Normal-hearing and hearing-impaired  groups are distinguished by color. Please zoom in for better view. }
    \label{fig:sentence_stats}
\end{figure*}

\subsection{2. Details of the UniCUE-HI Corpus}

To evaluate model effectiveness on hearing-impaired individuals, we construct the first large-scale Chinese Cued Speech (CS) dataset, UniCUE-HI Corpus, that includes both hearing-impaired and normal-hearing cuers. The proposed UniCUE-HI Corpus is designed to ensure linguistic and contextual diversity, covering a broad range of sentence categories, including daily communication scenarios, as illustrated in Figure~\ref{fig:dataset}.

To better understand linguistic variations across user groups, we analyze sentence structure statistics in our dataset. Figure~\ref{fig:sentence_stats} presents the distributions of sentence lengths and phoneme occurrences for normal-hearing and hearing-impaired cuers. While the overall distributions are similar, hearing-impaired users tend to produce slightly shorter sentences on average, indicating potential articulatory constraints and reduced fluency. These characteristics introduce additional modeling challenges and emphasize the importance of including hearing-impaired data for more realistic evaluation.

The dataset was constructed by recruiting 10 cuers (2 normal-hearing and 8 hearing-impaired), each instructed to simultaneously vocalize and encode text sentences sourced from~\cite{lei2024bridge}. All participants underwent systematic training to ensure accurate and fluent Mandarin Chinese CS production. In addition, we incorporate CS video data from 4 normal-hearing cuers in the MCCS dataset~\cite{lei2024bridge}, resulting in the UniCUE-HI Corpus with 14 cuers.

All data were collected with informed consent and are publicly released for open-source research purposes.

\begin{figure}[htbp] 
    \centering
    \includegraphics[width=1\linewidth]{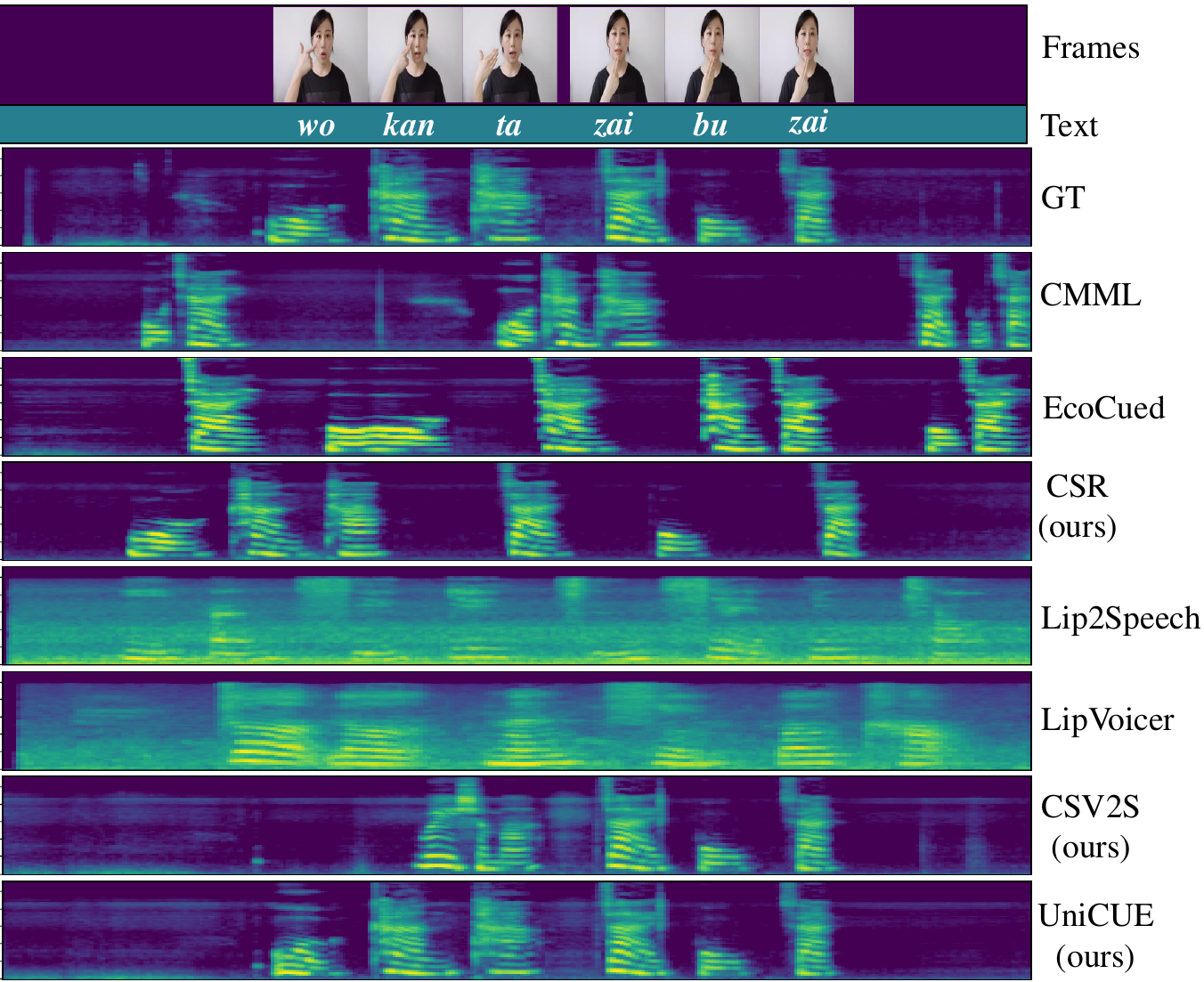}
    \caption{Visualization of the Mel-spectrograms of the generated speech. Our UniCUE model exhibits superior linguistic accuracy and temporal synchronization compared to baseline methods.}
    \label{fig:comparison}
\end{figure}

\subsection{3. Details of Comparison with CSR}

Since CSR methods output phoneme-level transcriptions, we design a two-stage pipeline for speech generation, as shown in Figure~\ref{fig:infer}(b).  
First, phoneme sequences are converted into Chinese characters using a rule-based CS mapping system, which follows a curated phoneme-to-character dictionary and handles tone-based disambiguation. To improve robustness and address homophonic ambiguities, the mapped text is further refined by the QWen language model\cite{qwen}.  
Then, the resulting character sequence is passed into F5-TTS~\cite{chen2024f5}, a controllable speech synthesizer that generates waveforms conditioned on reference speech from the same speaker. This design ensures a fair comparison with our direct CSV2S model in terms of speaker consistency and prosodic naturalness.

 \begin{table*}[htbp]
    \centering
    \caption{Ablation analysis the impact of hand cues in cued speech. (w/o) denotes using lip cues only, without hand gestures.
    \textbf{Bold} and \underline{underlined} results are the best and second-best results. $\uparrow$ indicates that larger values are better, while $\downarrow$ indicates that smaller values are preferable.  }
    \begin{tabular}{l|ccccc|cccc}
        \Xhline{1pt}  
        \multirow{2}{*}{Method} & \multicolumn{5}{c|}{Normal-hearing cuers} & \multicolumn{4}{c}{Hearing-impaired cuers} \\
        \cline{2-10}
        & WER $\downarrow$& LSE-C $\uparrow$ & LSE-D$\downarrow$ & DNSMOS $\uparrow$&STOI $\uparrow$&
        WER$\downarrow$& LSE-C$\uparrow$ & LSE-D$\downarrow$& DNSMOS $\uparrow$\\
        \Xhline{1pt}  
           GT & - & 7.274 & 7.314&2.79 &- & -&-&-&-\\
        \hline
         CSR(w/o)& 0.547&4.692&10.064&1.93&0.29&0.713&3.215&10.032&1.01\\
         CSR &\textbf{0.186} &4.874  & 9.125 &\textbf{2.53}&\textbf{0.57}&\textbf{0.224}& 3.342& 9.315&\textbf{2.29}\\  
         \hline
         CSV2S (w/o)&0.729&4.614&9.174&1.16&0.09&0.925&3.701&10.124&0.05\\
         CSV2S& 0.374&\underline{6.245}&\underline{7.962}&2.27&0.42&0.422&\underline{5.938}&\underline{8.347}&2.04\\
         \hline
         UniCUE (w/o)&0.592&4.934&8.927&1.88&0.26&0.793&4.412& 9.742&0.09\\
         UniCUE & \underline{0.205}& \textbf{6.729}& \textbf{7.632}&\underline{2.46}&\underline{0.53}&\underline{0.248} &\textbf{6.491}&\textbf{8.076}&\underline{2.17}\\
         \Xhline{1pt}  
    \end{tabular}
    \label{tab:hand_cues}
\end{table*}

\subsection{4. Qualitative Comparison with SOTA Methods}

To qualitatively assess the effectiveness of our proposed framework, we visualize the Mel-spectrograms of the generated speech and compare them against those produced by several state-of-the-art (SOTA) methods. As shown in Figure~\ref{fig:comparison}, UniCUE generates spectrograms that are more consistent with the ground truth in both temporal structure and phonetic detail. In particular, the generated patterns exhibit clearer formant trajectories and better alignment with visual speech cues, indicating superior temporal synchronization and articulation precision. These visual differences further support the quantitative gains observed in WER and LSE metrics, highlighting UniCUE’s ability to generate intelligible and temporally coherent speech from cued videos. 

\textbf{More generation results are provided in the supplementary demo file.}

\subsection{5. Discussion and Analysis}

\begin{figure*}[htbp]
    \centering
    \includegraphics[width=0.7\linewidth]{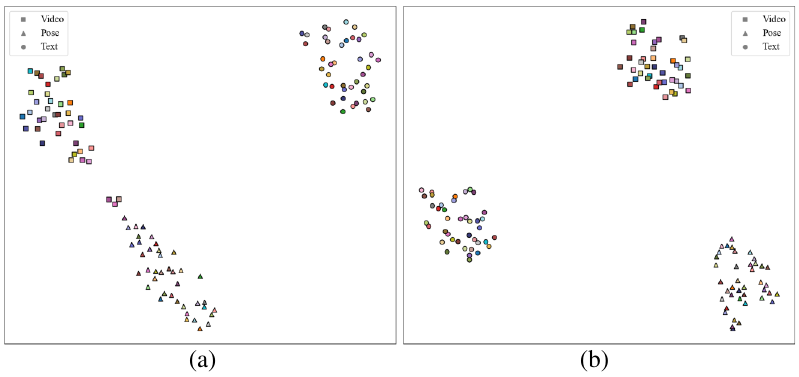}
    \caption{t-SNE visualization of video, pose, and text embeddings. (a): without Semantic Alignment Pool (SAP). (b): with SAP. Each color represents a different class, and different shapes denote different modalities. The results show that SAP enhances cross-modal alignment by bringing embeddings of the same class closer together in the semantic space.}
    \label{fig:sap}
\end{figure*}

\noindent\textbf{Visualization Analysis of Semantic Alignment Pool.}
To assess the effectiveness of the proposed Semantic Alignment Pool (SAP), we present t-SNE visualizations of the learned embeddings across video, pose, and text modalities in Figure~\ref{fig:sap}. Without SAP (Figure~\ref{fig:sap}a), the embeddings are loosely distributed with poor cross-modal alignment. In contrast, enabling SAP (Figure~\ref{fig:sap}b) results in compact and well-aligned clusters across modalities, indicating enhanced semantic coherence. These results demonstrate that SAP effectively bridges the visual-textual modality gap, facilitating grounded multimodal representation learning.

\noindent\textbf{Impact of Hand Cues.}  
Table~\ref{tab:hand_cues} presents an ablation study assessing the effect of hand cues in CS  speech synthesis.  
Removing hand gestures leads to substantial performance degradation across all metrics, particularly for hearing-impaired cuers.  
For instance, the WER of CSR increases drastically from 0.224 to 0.713 without hand cues, highlighting their crucial role in compensating for atypical articulatory patterns.  
Moreover, UniCUE achieves the best results in both LSE-C (6.729 for normal-hearing, 6.491 for hearing-impaired) and STOI (0.53), demonstrating its strong capability in generating expressive and intelligible speech by leveraging multi-modal visual information.

\begin{figure}[htbp]
    \centering
    \includegraphics[width=0.6\linewidth]{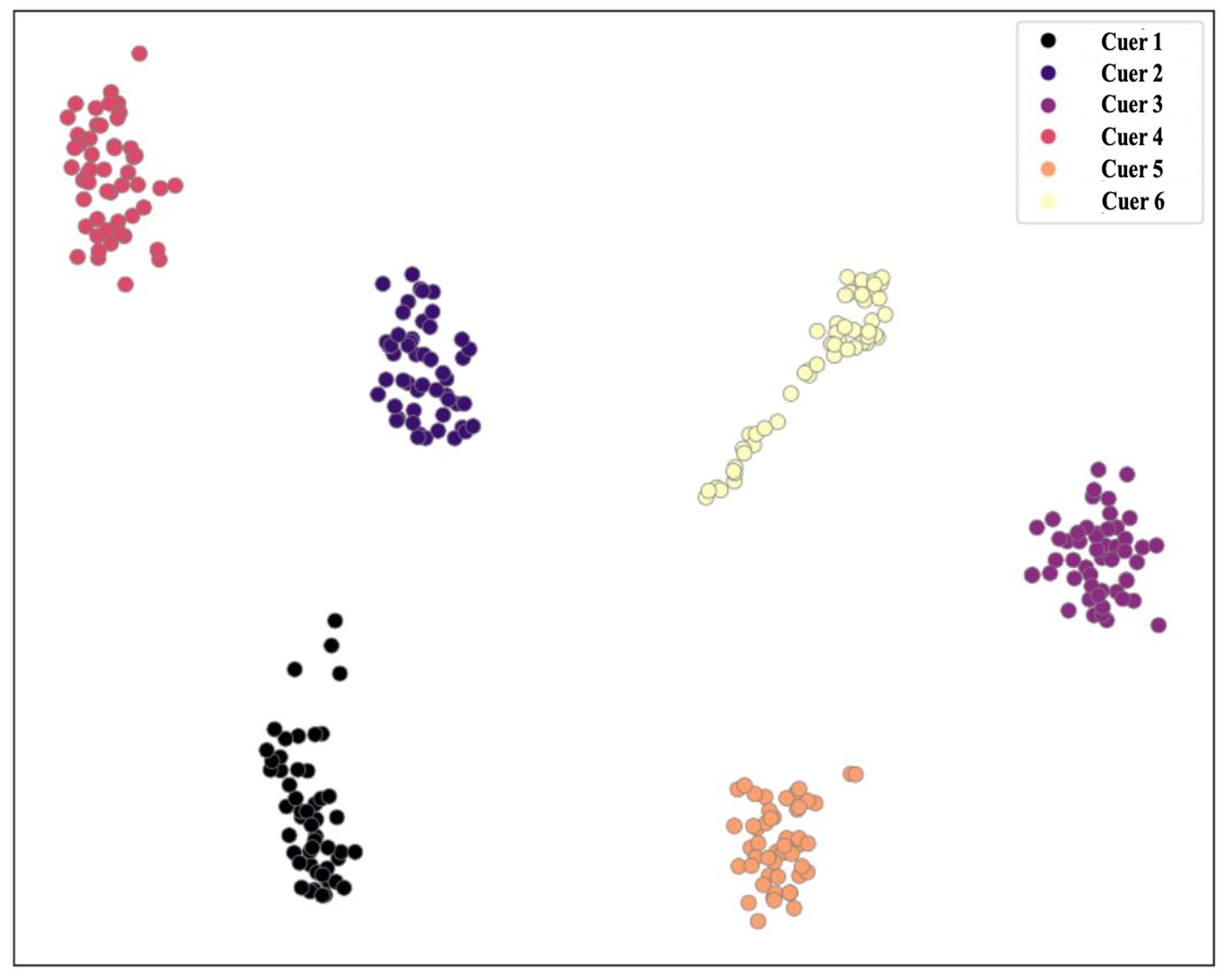}
    \caption{ The t-SNE visualization of 6 normal-hearing cuers embeddings from generated speech. Each point represents a speech sample. Better view by zooming in. }
    \label{fig:visual}
\end{figure}

\noindent\textbf{Analysis of Cuer Representation.}  
To quantitatively evaluate the model’s ability to preserve cuer-specific acoustic characteristics, we extract latent cuer embeddings from the generated speech using Resemblyzer \cite{resemblyzer2020}.  
These embeddings are projected into a 2D space via t-SNE \cite{maaten2008visualizing} for visualization.  
As illustrated in Figure~\ref{fig:visual}, speech samples from the same cuer (e.g., cuers 1 through 6) form compact clusters, while the distances between clusters corresponding to different cuers are notably larger.  
This demonstrates that the model effectively captures and differentiates personalized acoustic features unique to each cuer.

\begin{figure}[htbp]
    \centering
    \includegraphics[width=0.8\linewidth]{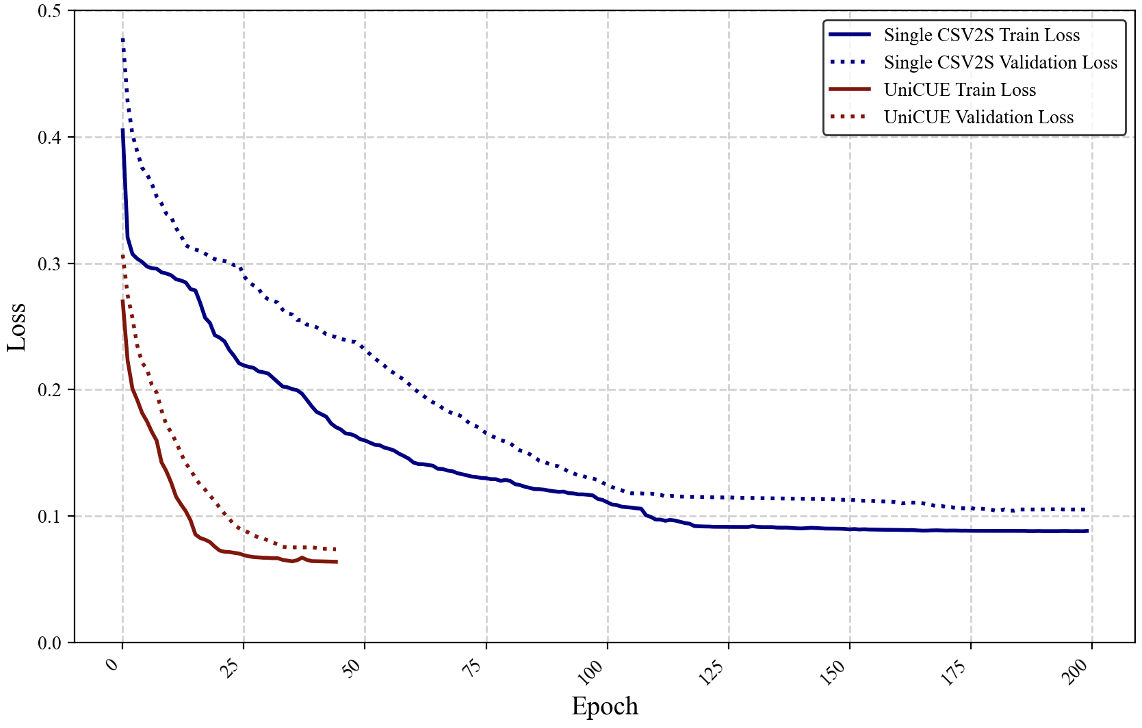}
    \caption{  
    The learning curves for the training and validation stages of our UniCUE model and the single CSV2S model.
 It can be seen that the efficiency and effectiveness of the unified approach. Better view by zooming in.}
    \label{fig:train_cost}
\end{figure}

\noindent\textbf{Analysis of Computational Efficiency.}  
To quantify the efficiency gains of our unified framework, we evaluate the time costs during both training and inference stages.  

\noindent\textbf{(a) Training Efficiency.}  
We compare the training performance of CSV2S and UniCUE. As shown in Figure~\ref{fig:train_cost}, UniCUE’s training loss converges faster to a stable value within fewer epochs, indicating that the incorporation of fine-grained visual understanding accelerates the learning process. Concurrently, the validation loss exhibits reduced fluctuation, reflecting enhanced model robustness.  

\noindent\textbf{(b) Real-time Inference.}  
To assess real-time capability, we compare the inference efficiency of the traditional pipeline (CSR + TTS) with our unified CSV2S framework. UniCUE achieves a significant reduction in inference time (4.15s), representing a 40\% speedup over the sequential CSR + TTS approach (6.93s), thereby demonstrating the efficiency benefits of the unified architecture.

\begin{table}[htbp]
    \centering
      \caption{Ablation results of loss design for CSR. \( \mathcal{L}_{\text{CE}}^{\text{masked}} \) denotes masked language modeling loss in CSR objective function. \textbf{H} and \textbf{HI} denote the normal-hearing data and hearing-impaired data, respectively.}
    \begin{tabular}{c|c|c}
    \hline
       Methods & WER$\downarrow$ (H)& WER$\downarrow$ (HI)\\
    \hline
      CSR (w/o $\mathcal{L}_{\text{CE}}^{\text{masked}}$)&0.214&0.263\\
       CSR & \textbf{0.186}&\textbf{0.224}\\
    \hline
       
    \end{tabular}
    
    \label{tab:csr_loss}
\end{table}

\subsection{6. Ablation Study of CSR Loss}

\noindent Table~\ref{tab:csr_loss} presents an ablation analysis of the hybrid loss design in our CSR module. Removing the masked language modeling loss \( \mathcal{L}_{\text{CE}}^{\text{masked}} \) leads to noticeable performance degradation, with WER increasing from 0.186 to 0.214 on normal-hearing data and from 0.224 to 0.263 on hearing-impaired data. This highlights the effectiveness of combining token-level masked supervision with sequence-level training, which enhances contextual understanding and improves transcription accuracy.

\end{document}